\documentclass[letterpaper]{article} 
\usepackage[]{aaai23}  
\usepackage{times}  
\usepackage{helvet}  
\usepackage{courier}  
\usepackage[hyphens]{url}  
\usepackage{graphicx} 
\usepackage{natbib}  
\usepackage{caption} 
\usepackage{algorithm}
\usepackage{algorithmic}
\usepackage{amssymb}
\usepackage{amsmath}
\usepackage{subfigure}
\usepackage[table]{xcolor}
\usepackage{multicol}
\usepackage{multirow}
\usepackage{comment}
\usepackage{tabularx}
\usepackage{booktabs}
\usepackage{utfsym}

\newcommand{\app}{\raise.17ex\hbox{$\scriptstyle\sim$}}

\usepackage{newfloat}
\usepackage{listings}
\DeclareCaptionStyle{ruled}{labelfont=normalfont,labelsep=colon,strut=off} 
\floatstyle{ruled}
\newfloat{listing}{tb}{lst}{}
\floatname{listing}{Listing}
\nocopyright 

\title{Generalizing Multiple Object Tracking to Unseen Domains by Introducing Natural Language Representation}
\author{
    En Yu\textsuperscript{\rm 1}\equalcontrib,
    Songtao Liu\textsuperscript{\rm 3}\equalcontrib,
    Zhuoling Li \textsuperscript{\rm 2},
    Jinrong Yang \textsuperscript{\rm 1},
    Zeming Li\textsuperscript{\rm 3},
    Shoudong Han\textsuperscript{\rm 1}\footnotemark[2],
    Wenbing Tao\textsuperscript{\rm 1}\thanks{Corresponding authors}
}
\affiliations{
    \textsuperscript{\rm 1}Huazhong University of Science and Technology,
    \textsuperscript{\rm 2}Tsinghua University,
    \textsuperscript{\rm 3}MEGVII Technology\\
    \{yuen, yangjinrong, shoudonghan, wenbingtao\}@hust.edu.cn}

\usepackage{bibentry}

\begin{document}

\maketitle

\begin{abstract}
Although existing multi-object tracking (MOT) algorithms have obtained competitive performance on various benchmarks, almost all of them train and validate models on the same domain. The domain generalization problem of MOT is hardly studied. To bridge this gap, we first draw the observation that the high-level information contained in natural language is domain invariant to different tracking domains. Based on this observation, we propose to introduce natural language representation into visual MOT models for boosting the domain generalization ability. However, it is infeasible to label every tracking target with a textual description. To tackle this problem, we design two modules, namely visual context prompting (VCP) and visual-language mixing (VLM). Specifically, VCP generates visual prompts based on the input frames. VLM joints the information in the generated visual prompts and the textual prompts from a pre-defined Trackbook to obtain instance-level pseudo textual description, which is domain invariant to different tracking scenes. Through training models on MOT17 and validating them on MOT20, we observe that the pseudo textual descriptions generated by our proposed modules improve the generalization performance of query-based trackers by large margins. To facilitate future research, we will release the code soon.

\end{abstract}

\section{Introduction}
\noindent As a fundamental vision perception task, multi-object tracking (MOT) has been extensively deployed in broad applications, e.g., autonomous driving, video analysis and intelligent robots \cite{bewley2016simple,wojke2017simple}. There exist numerous works studying how to track targets well \cite{tracktor,peng2020chained,pang2020tubetk,wang2019towards}. The early methods usually first detect targets using either anchor-based or keypoint-based detectors, and then associate the detected targets based on extracted appearance representation \cite{wang2020joint, zhang2021fairmot} or predicted motion \cite{tracktor,peng2020chained}. Recently, some researchers apply vision transformer \cite{dosovitskiy2020image, carion2020end} to MOT for implementing the models in a more end-to-end fashion \cite{meinhardt2021trackformer, zeng2021motr}.

\begin{figure}[t]
    \centering
    \subfigure[How the MOT17 dataset is split.]{
        \begin{minipage}{8.5cm} 
            \includegraphics[width=\textwidth]{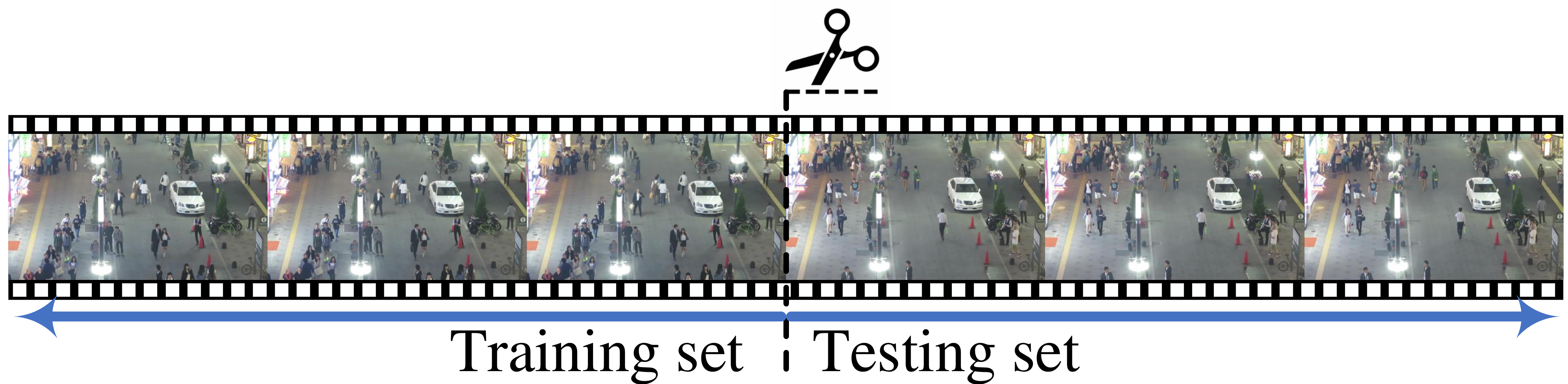} \\
        \end{minipage}
    }
    \subfigure[Cross-domain comparision]{
        \begin{minipage}{8.25cm}
            \includegraphics[width=\textwidth]{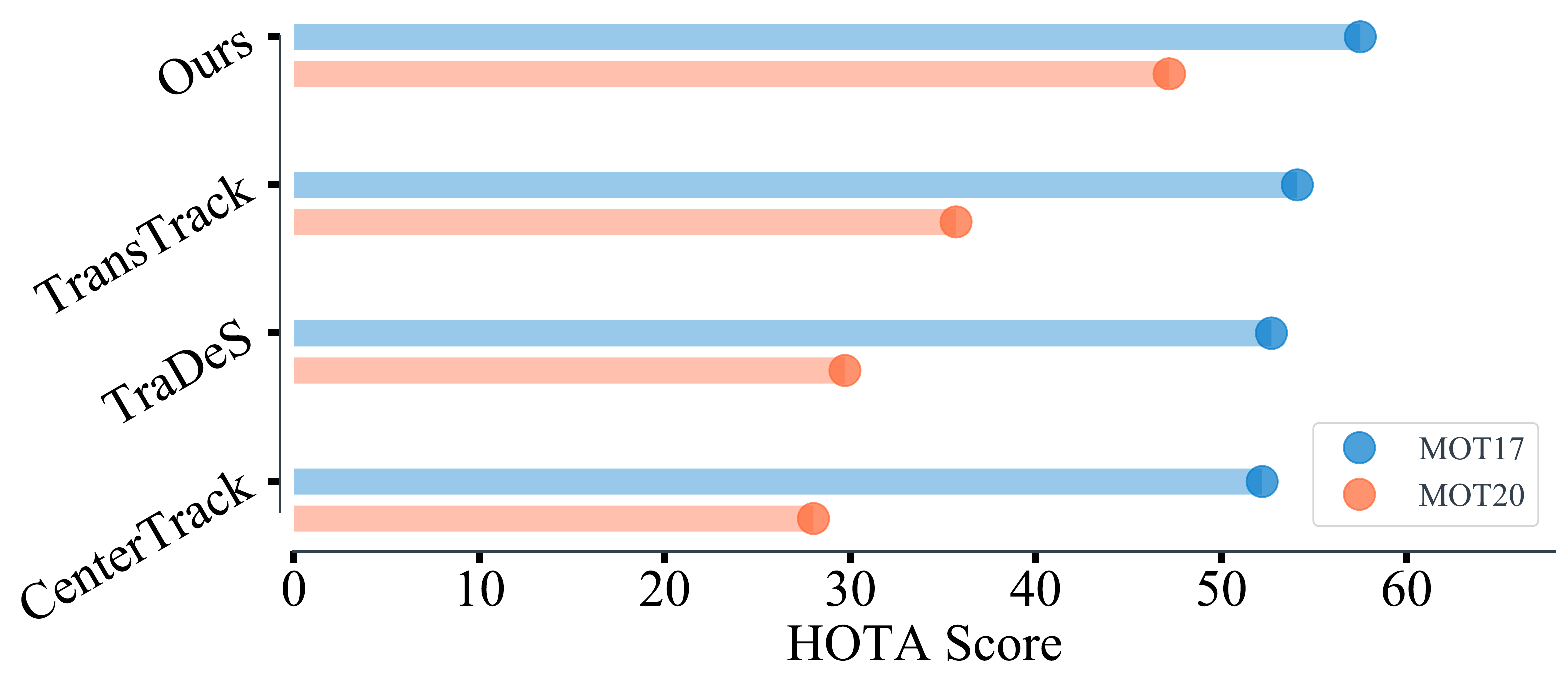} \\
            
        \end{minipage}
    }
\caption{(a) Since the training set and testing set of existing MOT datasets (such as MOT16 and MOT17) are usually created by splitting the same videos, they are completely from the same domains. (b) Compared with other counterparts, our method presents outstanding generalization ability when evaluated on an unseen domain.}
\label{fig1}
\end{figure}

Although previous MOT methods have achieved promising performance to some extent, almost all of them only consider training and evaluating models in the same domain. 
In fact, as shown in Fig. \ref{fig1} (a), the training and testing sets of commonly employed MOT datasets, such as MOT17 and MOT20, are mostly produced by spliting the identical videos into two parts.
Thus the existing models are only evaluated in the same domain as the training set. When we test some of them (i.e., TransTrack, TraDeS, and CenterTrack) in another unseen domain, as illustrated in Fig. \ref{fig1} (b), their performance drops dramastically, leading to generalization bottleneck. And to our best knowledge, this domain generalization problem of MOT is hardly studied in literature.

\begin{figure}[t]
\centering
\includegraphics[width=0.9\columnwidth]{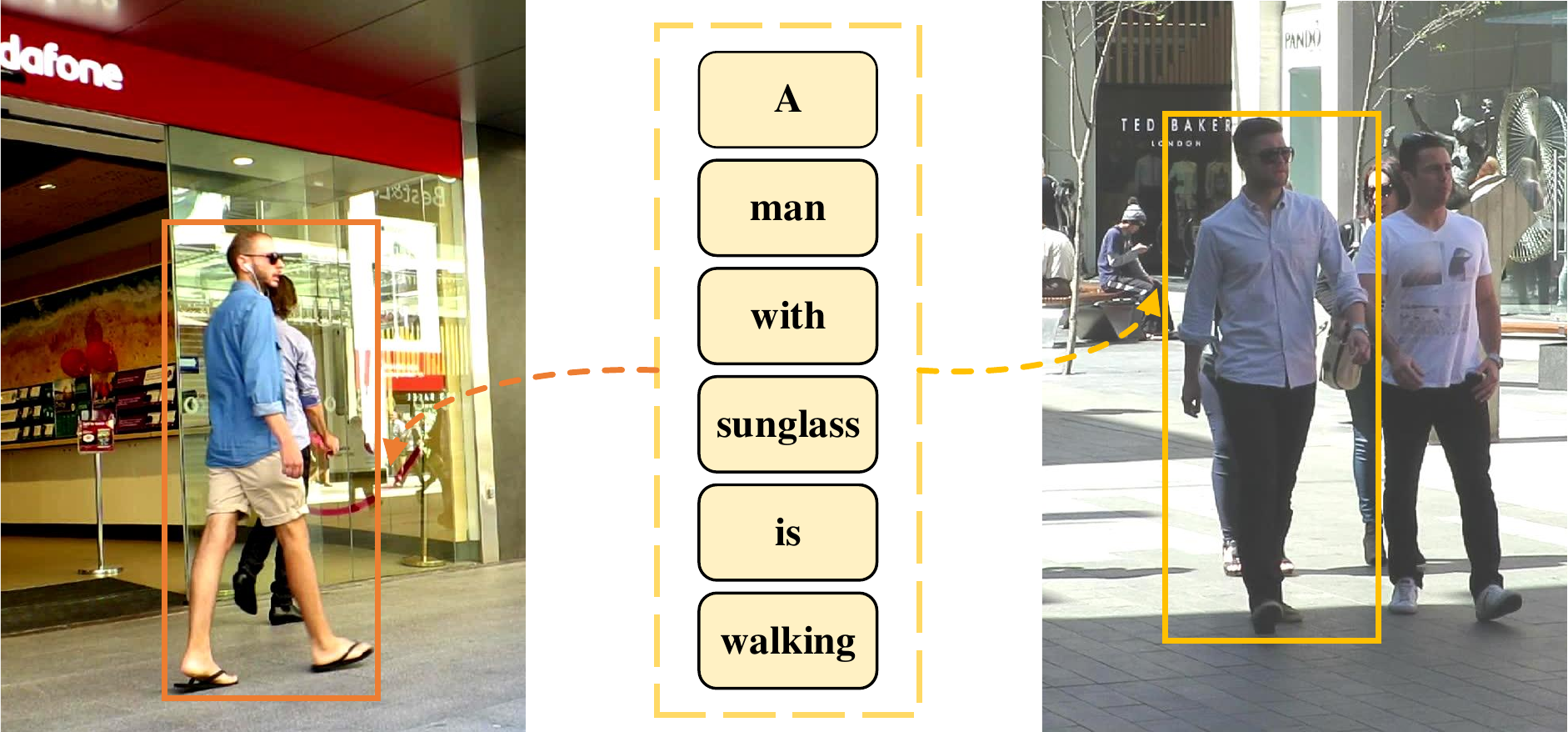} 
\caption{\textbf{The generalization ability of natural language representation.} As shown, since the information in a sentence is high-level, it can describe persons with similar appearances in two significantly different domains.}
\label{fig:language}
\label{fig2}
\end{figure}

In this work, we aim to tackle the poor generalization ability of exist trackers where the cross-domain training data is unattainable. 
Inspired by the recent works of multi-modal models, we try to seek more generalized features besides vision.
To this end, we first draw the observation that natural language description is naturally domain invariant to different tracking domains. For example,  as illustrated in Fig. \ref{fig2}, although the two men appear in two significantly different domains (e.g., illumination, background, or resolution), they can be described by the same sentence. Hence, we believe that natural language representation generalizes better. Moreover, the recent models like CLIP \cite{radford2021learning} provide unified visual and linguistic representations, allowing us to design some advanced strategies to introduce the natural language representation into MOT models.


To achieve this goal, the key challenge is how to design a model for adaptively pre-processing and effectively post-aggregating both natural language and images. CLIP-like models only bridge the features from two modals, while our model should learn to improve its generalization ability in the tracking task. By virtue of the versatile multi-modal fusion property (e.g., image, text, or point cloud) of the Transformer, we select a Transformer-based tracker, MOTR \cite{zeng2021motr}, as the baseline model. In MOTR, every tracklet is represented as a track query. Hence, it is straightforward to devise modules for incorporating the text representation into the original track queries of MOTR. 

However, there is still a severe obstacle hindering our design: it is too troublesome to label every tracking target in the training set with description text. To address this problem, we first prepare a Trackbook, which comprises 56 description phrases covering most tracking cases. In this way, the following problem is how to associate the information in these description phrases with detected targets. 

To this end, we first convert the phrases in Trackbook to word embeddings named textual prompts through bag-of-words \cite{zhang2010understanding}. Afterwards, we design two modules i.e., Visual Context Prompting (VCP) and Vision-Language Mixture (VLM), to associate the description text in the Trackbook with tracked targets automatically. VCP encodes the input image contexts as tokens called visual prompts, which contain domain related appearance and scene information of tracking targets. This information can be regarded as the implicit description of targets and used as the supplement to the domain-independent textual prompts. VLM associates and mixes the aforementioned textual prompts and visual prompts for generating new embeddings called pseudo textual description (PTD). The PTD contains high-level descriptions about tracking targets and generalizes well like the sentences presented in Fig. \ref{fig2}. Therefore, by combining the pseudo textual descriptions with the original track queries, the generalization ability of the baseline tracker is improved. 

Through combining the proposed method of generating PTD with the baseline method, we obtain a new tracking model and name it as \textit{language-guided tracker} (LTrack). To evaluate its generalization performance and compare with other advanced methods, we train all models on MOT17 and validate on MOT20 dataset. To our knowledge, this is the first work to leverage this testing protocal, which is exactly a new cross-domain MOT evaluation benchmark. The experimental results reveal that our method boosts the generalization ability of the baseline model significantly while achieving state-of-the-art (SOTA) performance on this benchmark.

\begin{figure*}[t]
\centering
\includegraphics[width=1.6\columnwidth]{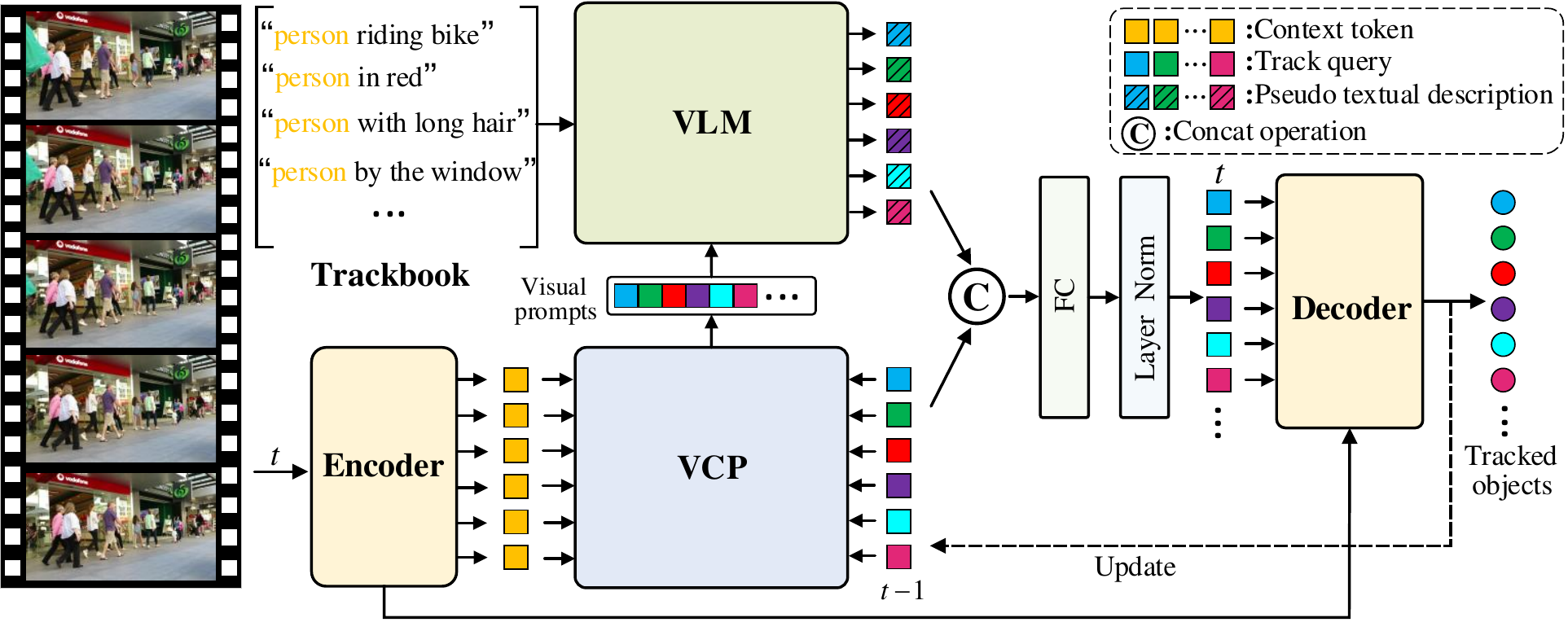} 
\caption{\textbf{Overall pipeline of LTrack}.}
\label{fig3}
\end{figure*}

\section{Related Work}
\noindent \textbf{Multiple-object Tracking.} Thanks to the fast development of object detection techniques \cite{ren2015faster, tian2019fcos, zhou2019objects, carion2020end}, existing trackers mainly follow the tracking-by-detection paradigm \cite{bewley2016simple, wojke2017simple, tracktor, peng2020chained, pang2020tubetk, wang2019towards}, which first localizes targets in each frame and then associates them based on their recognized identities to obtain trajectories.

According to the association strategy, early MOT methods can be further divided into motion-based trackers and appearance-based trackers. Most motion-based trackers perform the association step based on motion prediction algorithms, such as Kalman Filter \cite{welch1995introduction} and optical flow \cite{baker2004lucas}. There are also some other motion-based trackers \cite{feichtenhofer2017detect, tracktor, CenterTrack, sun2020transtrack, shuai2021siammot} that build networks to directly predict the future locations or displacements of concerned targets. In contrast to the motion-based trackers, the appearance-based trackers \cite{wojke2017simple, he2021learnable, wang2019towards, zhang2021fairmot, liang2020rethinking, wu2021track, yu2022relationtrack} first extract the appearance representation of targets and match targets to trajectories based on the similarity of the obtained representation.

Although the performance of the aforementioned methods is competitive, they do not realize fully end-to-end tracking due to the post-processing operations. Recently, The Transformer model originally designed for natural language processing (NLP) has been applied to computer vision. For example, DETR \cite{carion2020end} models the 2D object detection task as a set-to-set prediction problem based on Transformer \cite{vaswani2017attention}. DETR \cite{carion2020end} is first proposed to turn the object detection into a set prediction problem. Inspired by DETR, TrackFormer \cite{meinhardt2021trackformer} and MOTR \cite{zeng2021motr} regard MOT as a sequence prediction problem by representing every trajectory as a track query. They are fully end-to-end and do not demand post-processing. In this work, MOTR is taken as the baseline method.

\noindent \textbf{Vision-language Models.} Vision-language  models have been widely studied in the fields of text-to-image retrieval \cite{wang2019camp}, images caption \cite{xu2015show}, visual question answering \cite{antol2015vqa}, referring segmentation \cite{hu2016segmentation}, etc. Among the many related publications, vision-language pretraining has attracted growing attention recently, and the milestone work is Contrastive Language-Image Pre-training (CLIP) \cite{radford2021learning}. CLIP pre-trains models through conducting contrastive learning among 400 million image-text pairs crawled from the Internet. Its impressive generalization ability has been confirmed by being evaluated across 30 classification datasets. The pre-trained CLIP encoders are also applied to many other downstream tasks, e.g., open-vocabulary detection \cite{gu2021open}, zero-shot semantic segmentation \cite{rao2021denseclip} and so on. However, to the best as we know, CLIP encoders have not been applied to MOT, and we are the first to utilize the knowledge contained in the CLIP encoders to boost the generalization performance of visual MOT models. 

\noindent \textbf{Domain Generalization for MOT.} Domain generalization is critical for practical applications of visual deep learning models. When evaluated on an unseen testing domain, a model is expected to present performance consistent with that of the training domain. Although the domain generalization problem has been extensively studied in many tasks, such as object detection \cite{chen2018domain} and person re-identification \cite{deng2018image}, it is hardly studied in MOT. In this work, we aim to bridge this gap.

\section{Methodology}

\subsection{Delve into the Track Query}
\label{motr}

MOTR is an end-to-end MOT tracker that is build on the transformer. The key design of MOTR is the track query. In fact, MOTR employs two kinds of learnable embeddings, i.e., the detect queries and track queries. The detect queries are used in the same way as DETR \cite{carion2020end}, and the difference is that they are only responsible for recognizing newborn targets in each frame rather than all targets. When a new object is detected, the matched detect query obtains a corresponding track query to track this target. Meanwhile, if a previously detected object is continuously missing for $N_{m}$ frames ($N_{m}$ is a hyper-parameter), the track query is deleted. Hence, the track query naturally represents a trajectory, and no post-processing such as NMS is demanded. 

The track queries are continuously updated according to the representation extracted from the input frame during the tracking process, Nevertheless, the representation is only learned from low-level image features and represents the statistics of the training set. Hence, they are prone to being over-fitting to a specific tracking domain. When the tracking domain changes, the tracker could fail. To tackle this problem, we hope to introduce high-level natural language representation into track queries to boost the generalization capability.

\subsection{Method Overview} 

As illustrated in Fig. \ref{fig3}, LTrack comprises four components, the original MOTR (including the shown encoders and decoders). the Trackbook, and the proposed VCP and VLM modules. Given the $t$ frame of a video as input, the encoders extract context tokens from it. The context tokens and the track queries of the $t-1$ frame are fed to VCP to produce visual prompts. Afterwards, the visual prompts and the textual prompts generated from the Trackbook are transformed as PTD tensors by VLM. The CLIP text encoder is employed in VLM to transfer the prompts to the PTD. Finally, the track queries of the $t-1$ frame and the PTD tensors are concatenated and encoded as the track queries of the $t$ frame for further tracking.

\subsection{Trackbook}
\label{Trackbook}

As mentioned before, it is too troublesome to label every tracking target with a descriptive sentence. Alternatively, we create a dictionary called Trackbook including 56 description phrases, such as "person riding bike". These phrases can describe most tracking cases. Then, we transform these phrases into more intact sentences by adding a fixed template. For example, "person riding bike" is converted as "A photo of person riding bike". Afterwards, we convert all these sentences as tensors named textual prompts through the classical bag-of-words method \cite{zhang2010understanding}. The textual prompts contain the description information of various kinds of tracking targets. We hope our designed modules can learn to associate a target with the corresponding description information in the textual prompts automatically and use the information to improve the generalization performance. Notably, the Trackbook can be dynamically supplemented. And the learned PTD can be more general when applying more detailed textual descriptions in the Trackbook. 

\subsection{Visual Context Prompting}  

We aim to associate the textual information in Trackbook with the visual information of tracking targets. In addition, we hope these two kinds of information can be processed by a unified module. Thus, we hope the visual and textual information can be organized in the same format (such as tensors of the same dimension). Since textual information is often represented as prompts, the visual information should also be modeled as prompts. In this way, we design a module to learn the association between them later. Moreover, the textual information from the Trackbook is coarse-grained and the number of texts is limited. Therefore, we need to capture the fine-grained information from the image contexts to enrich the textual description.

To the aforementioned end, we design the VCP module. The VCP module is for generating visual prompts based on the context tokens generated by the encoders shown in Fig. \ref{fig3}. To clearly inform VCP which regions it should focus on, the track queries of the previous frame are also adopted as the input of VCP. Specifically, VCP adopts the Transformer decoder structure. It takes the context tokens produced by the final encoder (denoted as $c_{t}$) and the track queries of the previous frame $q_{t-1}$ as input, which is formulated as follows:
\begin{equation}
v_{t} = TransDecoder(q_{t-1}, c_{t}),
  \label{Eq1}
\end{equation}
where $v_{t} \in R^{M \times D}$ denotes the obtained visual prompts. $M$ and $D$ are the track query number and the embedding dimension, respectively. In VCP, the Query of the Transformer decoder is obtained based on the original track query $q_{t-1}$, and the Key and Value are derived from $c_{t}$.

\subsection{Visual-Language Mixing} 

\begin{figure}[t]
\centering
\includegraphics[width=0.62\columnwidth]{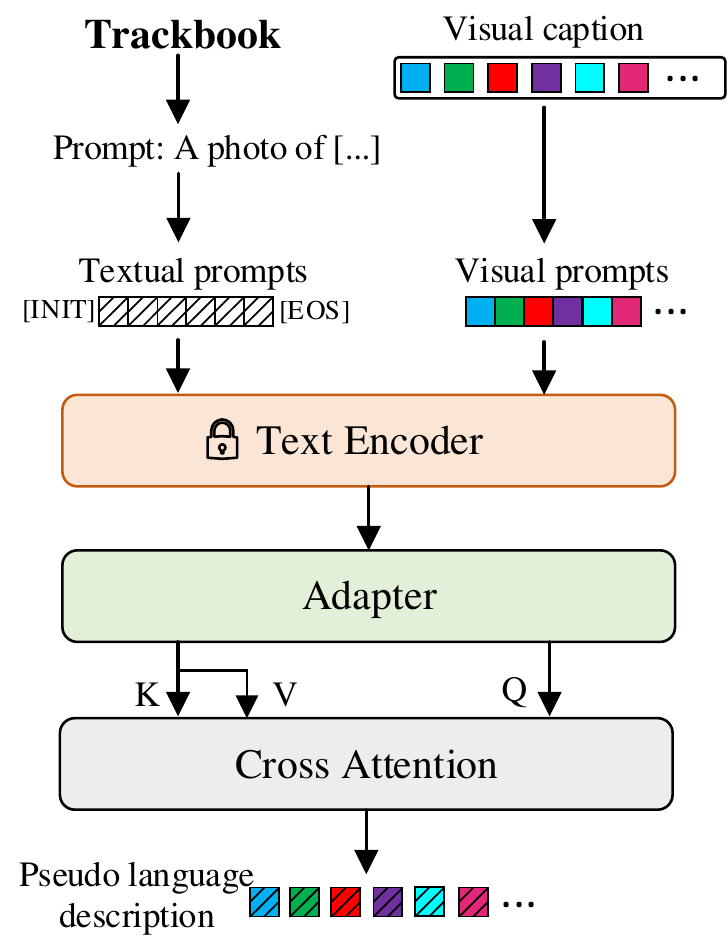} 
\caption{\textbf{Illustration of the visual-language mixing (VLM) module,}  which consists of three components: (1) A pre-trained CLIP text encoder ($[INIT]$ and $[EOS]$ are the tokens to mark the begin and end of the text sequence). (2) An adapter module. (3) A cross-attention module.}
\label{fig4}
\end{figure}

VLM is responsible for mixing the textual prompts from Trackbook and the visual prompts from VCP to obtain PTD. The detailed structure of VLM is illustrated in Fig. \ref{fig4}. As shown, VLM consists of three parts, i.e., the text encoder, the adapter, and a cross-attention module. The text encoder is exactly the publically available CLIP text encoder, which is pre-trained using about 400 million image-text pairs. In this work, the weights of this text encoder are fixed and not updated during the training process in order to protect the original CLIP knowledge.

Since the weights of the text encoder are fixed, we design an adapter network after this text encoder to blend new knowledge with the original CLIP. The adapter is a MLP block with a residual connection. Mathematically, it can be formulated as:

\begin{equation}
T_a(x) = Relu(T(x)^{T}W_1)W_2 + x,
  \label{Eq2}
\end{equation}
where $W_1$ and $W_2$ are linear transformation weights, and $T(\cdot)$ denotes the CLIP text encoder. Notably, in the text encoder and adapter modules, the textual and visual prompts are processed independently. 

After the adapter module, a cross-attention module is built to associate the textual representation and visual representation. It also adopts the Transformer decoder structure, which can be formulated as:

\begin{equation}
\begin{aligned}
&v = [v_1,v_2,..., v_M], \\
&t = [t_1,t_2,...,t_K], \\
&l = CrossAttn(T_{a}(v), T_{a}(t)),
  \label{Eq2}
\end{aligned}
\end{equation}
where $M$ represents the visual prompt length and $K$ denotes the textual prompt length. In this work, we only employ the CLIP text encoder and do not use the CLIP image encoder. It is because we observe that the  representation produced by the CLIP image encoder is unsuitable for tasks including localization.

The output of the cross-attention module in VLM is the learned PTD. By concatenating it with the track queries of the previous frame and further transforming the concatenated tensors (the FC and layer normalization in Fig. \ref{fig3}), we obtain the track queries of the t frame. These enhanced track queries are used to update the tracklets using the algorithm proposed by MOTR, leading to promising performance.

\subsection{Optimization}

Given a clip $V_{\xi}$ of $N$ frames as input, the results predicted by the model are denoted as $\widehat{P}=\{\hat{p}_{i}\}_{i=1}^{N}$, and the corresponding ground-truths are $P=\{p_{i}\}_{i=1}^{N}$. The overall loss $L_{clip}$ is computed based on $\widehat{P}$ and $P$. It consists of two parts, the tracking loss and detection loss. These two losses exactly share the same form. The difference is that the tracking loss is for localizing the targets that have been recognized in previous frames, and the detection loss is to tackle the newborn targets. Mathematically, $L_{clip}$ can be formulated as follows:
\begin{equation}
  \begin{split}
    \hspace{-3mm}\mathcal{L}_{clip}\hspace{-0.5mm}=\hspace{-0.5mm} \frac{\sum\limits_{n=1}^{N}\hspace{-0.5mm}(\mathcal{L}(\widehat{P}^{i}_{tr}|_{q_{t}},P^{i}_{tr})\hspace{-0.8mm}+\hspace{-0.8mm}\mathcal{L}(\widehat{P}^{i}_{det}|_{q_{d}},P^{i}_{det})\hspace{-0.5mm}) }{\sum\limits_{n=1}^{N}(T_{i})},
  \end{split}
  \label{Eq5}
\end{equation}
where $\widehat{P}^{i}_{tr}|_{q_{t}}$, $\widehat{P}^{i}_{tr}$, $\widehat{P}^{i}_{det}|_{q_{d}}$, and $\widehat{P}^{i}_{det}|_{q_{d}}$ are the association predictions, association labels, detection predictions, and prediction labels, respectively.  $T_i$ denotes the total number of the targets in the $i-{\rm th}$ frame. $\mathcal{L}(\cdot)$ is implemented similarily to the one in DETR, which is formulated as:
\begin{equation}
  \begin{split}
    \mathcal{L}(\widehat{P}_{i}|_{q_{i}},P_{i}) = \lambda_{cls}\mathcal{L}_{cls} + \lambda_{l_{1}}\mathcal{L}_{l_{1}} + \lambda_{giou}\mathcal{L}_{giou},
  \end{split}
  \label{Eq6}
\end{equation}
where $\mathcal{L}_{cls}$,  $\mathcal{L}_{l_{1}}$, and $\mathcal{L}_{giou}$ are the focal loss \cite{lin2017focal} for classification, $L_{1}$ loss for regressing width and height, and the common generalized IoU loss. $\lambda_{cls}$, $\lambda_{l_1}$, $\lambda_{giou}$ are three hyper-parameters.

\begin{table*}[!t]
\begin{center}
\resizebox{17cm}{4.2cm}{
\begin{tabular}{ccccccccc} \toprule
Method & Published & Data & HOTA$\uparrow$ & AssA$\uparrow$ & DetA$\uparrow$ & MOTA$\uparrow$ & IDF1$\uparrow$ & IDS$\downarrow$ \\ \hline

\midrule  
\multicolumn{9}{c}{\bf{MOT17 (In-domain)}}\\
\midrule  
TubeTK$^*$~\cite{pang2020tubetk}      & CVPR20 & 17 & 48.0 & 45.1 & 51.4 & 63.0 & 58.6 & 4317 \\
CTracker$^*$~\cite{peng2020chained}   & ECCV20 & 17 & 49.0 & 45.2 & 53.6 & 66.6 & 57.4 & 5529 \\
QDTrack$^*$~\cite{pang2021quasi} & CVPR21 & 17 & 53.9 & 52.7 & 55.6 & 68.7 & 66.3 & 3378 \\
FairMOT$^*$~\cite{zhang2021fairmot}    & IJCV21 & CH+EX+17 & 59.3 & 58.0 & 60.9 & 73.7 & 72.3 & 3303 \\
CSTrack$^*$~\cite{liang2020rethinking}    & Arxiv20 & EX+17  & - & - & - & 70.6 & 71.6 & 3465 \\
CenterTrack~\cite{CenterTrack} &ECCV20 & CH+17  & 52.2 & 51.0 & 53.8 & 67.8 & 64.7 & 3039 \\
TraDeS~\cite{wu2021track}       & CVPR21 & CH+17 & 52.7 & 50.8 & 55.2 & 69.1 & 63.9 & 3555 \\
TransTrack~\cite{sun2020transtrack} & Arxiv20 & CH+17 & 54.1 & 47.9 & \bf{61.6} & \bf{74.5} & 63.9 & 3663 \\
TransCenter~\cite{xu2021transcenter} & Arxiv21 & CH+17 & 54.5 & 49.7 & - & 73.2 & 62.2 & 3663 \\
MOTR~\cite{zeng2021motr} & ECCV22 & CH+17 & 57.2 & 55.8 & 58.9 & 71.9 & 68.4 & 2115 \\
\bf{LTrack}                          & Ours & CH+17  & \bf{57.5} & \bf{56.1} & 59.4 & 72.1 & \bf{69.1} & \bf{2100} \\
\midrule  %
\multicolumn{9}{c}{\bf{MOT20 (Cross-domain)}}\\
\midrule  
CenterTrack~\cite{CenterTrack} &ECCV20 & CH+17  & 29.7 & 25.6 & 35.0 & 42.9 & 39.0 & 6397 \\
FairMOT~\cite{zhang2021fairmot}    & IJCV21 & CH+17 & 41.9 & 35.9 & \bf{49.7} & 57.6 & 53.8 & 13621  \\
TraDeS~\cite{wu2021track}       & CVPR21 & CH+17 & 28.0 & 25.5 & 32.7 & 44.9 & 39.3 & 7729 \\
CSTrack~\cite{liang2020rethinking}    & Arxiv20 & CH+17  & 33.9 & 29.8 & 38.8 & 49.6 & 44.9 & 10041 \\
TransTrack~\cite{sun2020transtrack} & Arxiv20 & CH+17 & 35.8 & 27.3 & 47.3 & \bf{58.1} & 44.8 & 13189 \\
OMC~\cite{liang2021one} & AAAI22 & CH+17 & 38.8 & 32.2 & 46.9 & 55.9 & 49.4 & 13813 \\
MTrack~\cite{yu2022towards} & CVPR22 & CH+17 & 40.6 & 37.0 & 44.9 & 54.8 & 52.9 & 7639 \\
MOTR~\cite{zeng2021motr} & ECCV22 & CH+17 & 43.7 & 41.7 & 45.9 & 56.0 & 56.8 & 2184 \\
\bf{LTrack }                           &    Ours  & CH+17  & \bf{46.8} & \bf{45.4} & 48.4 & 57.8 & \bf{61.1} & \bf{1841} \\
\bottomrule
\end{tabular}}
\end{center}
\caption{\textbf{Performance comparison with preceding SOTAs on the in-domain and cross-domain evaluation benchmarks.} ↑/↓ indicates that higher/lower score is better. Previous MOT trackers are often trained using different data volumes. For the in-domain evaluation part (MOT17) of this table, $^*$ means different training data setting is used. The used datasets of all methods are marked out in the column ``Data'' of this table (CH, 17, and EX refer to CrowdHuman, MOT17 and more extra data, respectively). In the cross-domain evaluation part, all methods are trained and validated in the same data setting.}
\label{tab1}%
\end{table*}

\section{Experiments}

\subsection{Benchmark} 
As mentioned before, almost all previous works train and validate their trackers in the same domain, and the available datasets are usually collected in a single domain. However, in this work, we want to evaluate the domain generalization ability of trackers in unseen domains. Hence, we need to design a new cross-domain benchmark. To this end, we first revisit existing MOT datasets carefully, and select MOT17 \cite{milan2016mot16} and MOT20 \cite{dendorfer2020mot20} to build this benchmark. In the following, we first introduce MOT17 and MOT20 separately. Afterwards, we explain how we use them to compose the cross-domain benchmark. Finally, the used metrics are described.

\noindent \textbf{MOT17 and MOT20.} The MOT17 dataset consists of 14 video sequences. Among them, 7 sequences are for training and the other 7 sequences are used to validate models. These sequences cover various scenarios and weathers. Compared with MOT17, MOT20 is more challenging. It is composed of 8 long video sequences captured in 3 crowded scenes. In some scenes, more than 220 pedestrians are contained in a single frame. Besides, the scenes in the MOT20 dataset are more diversified including indoor and outdoor, day and night, etc.

\noindent \textbf{Cross-domain Benchmark.} To verify the domain generalization ability of models in unseen domains, we use MOT17 to train models and validate them in MOT20. Since there exists a significant domain gap between MOT17 and MOT20, this evaluation setting can serve as a new cross-domain benchmark. The adopted metrics in this benchmark include the HOTA \cite{hota2021} and CLEAR-MOT Metrics \cite{bernardin2008evaluating}. Specifically, HOTA consists of higher order tracking accuracy (HOTA), association accuracy score (AssA) and detection accuracy score (DetA). CLEAR-MOT Metrics include ID F1 score (IDF1), multiple object tracking accuracy (MOTA) and identity switches (IDS). Among them, HOTA, AssA, IDF1 are the most important metrics for comparing tracking performance. And all the metrics are evaluated by TrackEval \footnote{TrackEval: \url{https://github.com/JonathonLuiten/TrackEval}}.

\subsection{Implementation Details}

Following MOTR, we build LTrack based on Deformable-DETR \cite{zhu2020deformdetr}, which is pre-trained on COCO \cite{coco} and employs ResNet50 \cite{He2016Resnet} as the image encoder. During the training process, the batch size is 1 and each batch contains a multi-frame video clip. The length of each video clip is $2$ at the beginning of training, and it is increased by $1$ after every $50$ epochs. The frames in each clip are selected from training videos with a  random interval between $1$ to $10$. We employ Adam as the optimizer \cite{kingma2014adam} and the initial learning rate is set to $2 \times 10^{-4}$. LTrack is trained for totally 200 epochs and the learning rate decays by $10$ at the $100{\rm th}$ epoch. The data augmentation strategy of LTrack follows the setting in MOTR, which includes random flip and random crop. During the training process, $\lambda_{cls}$, $\lambda_{l_1}$, and $\lambda_{giou}$ are set as $2$, $5$, and $2$, respectively.

\subsection{Comparison with previous SOTA Methods}

In this part, we compare the performance of LTrack with preceding SOTA methods. The comparison is conducted under two protocols, in-domain evaluation and cross-domain evaluation. For in-domain evaluation, all models are trained using the MOT17 training set and the CrowdHuman dataset (a 2D pedestrian dataset) \cite{shao2018CrowdHuman}, and the trained models are verified with the MOT17 validation set. Notably, incorporating CrowdHuman into the training process is a common operation in previous publications. Thus, we also adopt it. For the in-domain evaluation part of Tab. \ref{tab1}, the methods marked in gray are trained in the same setting as LTrack. In the cross-domain evaluation protocol, all the methods are trained using the same data setting, and they are tested in the MOT20 validation set. The evaluation results are reported in Tab. \ref{tab1}.

\noindent \textbf{In-domain Evaluation.} As illustrated in Tab. \ref{tab1}, LTrack obtains the metrics HOTA of $57.5\%$, AssA of $56.1\%$, and IDF1 of $69.1\%$ on MOT17. This means LTrack outperforms all compared methods in the in-domain evaluation setting. The results indicate that LTrack can tackle in-domain tracking scenes well. Notably, LTrack also obtains promising performance on the detection related metrics ($59.4\%$ DetA and $72.1\%$ MOTA). Meanwhile, it only produces $2100$ IDS, which is the lowest among all the methods. Therefore, the tracklets generated by LTrack are more continuous.

\noindent \textbf{Cross-domain Evaluation.} As presented in Tab. \ref{tab1}, when we test the compared methods in our built cross-domain evaluation benchmark, their performance drops significantly. By contrast, the performance drop of LTrack is relatively small. For instance, in MOT20, LTrack achieves $46.8\%$ HOTA, $45.4\%$ AssA and $61,1\%$ IDF1, which significantly outperforms all compared methods by large margins. The results indicate that the generalization ability of LTrack to unseen domains is promising. In addition, LTrack surpasses its baseline method, MOTR, by $3.1\%$ ($46.8\%$ vs. $43.7\%$) on HOTA, $3.7\%$ ($45.4\%$ vs. $41.7\%$) on AssA and $4.3\%$ ($61.1\%$ vs. $56.8\%$) on IDF1, which further confirms the benefit of introducing text representation into visual MOT.

\subsection{Ablation Study}

In this subsection, we verify the effectiveness of the proposed modules separately through ablation studies. All the experiments are conducted on the cross-domain evaluation benchmark. To accelerate the ablation study process, We use a lite version of LTrack by reducing the number of Transformer encoders from 6 to 1. The models are trained on the MOT17 training set for 200 epochs.

\noindent \textbf{Analysis on VCP.} In this part, we conduct an in-depth analysis of VCP. VCP adopts the transformer decoder structure to produce visual prompts. Among its inputs, the information in visual context tokens is crucial. However, we are unclear about which visual features from previous modules (the backbone and Transformer encoders) should be taken as the context tokens. To address this problem, We select the last three stages' feature maps of the backbone ResNet-50 and the output of the transformer encoder as the input to VCP separately. The results are reported in Tab. \ref{tab2}.

\begin{table}[t]
  \centering
    \scalebox{0.75}{
\small
  \begin{tabular}{@{}c ccccc@{}}
    \toprule
     Key feature & HOTA$\uparrow$ & AssA$\uparrow$ & MOTA$\uparrow$ & IDF1$\uparrow$ & IDS$\downarrow$ \\
    \midrule
    c3           & 15.8 & 19.9 & 12.0 & 19.0 & 5157 \\
    c4            & 18.3 & 21.1 & 14.5 & 22.1 & 4849 \\
    c5            & 15.5 & 16.6 & 15.0 & 19.1 & 6712 \\
    \midrule
    \textbf{Enc}   & \bf 19.2 & \bf 21.9 & \bf 16.7 & \bf 24.1 & \bf 4674 \\
    \bottomrule
  \end{tabular}}
  \caption{Comparison between different visual features to VCP. c3, c4, and c5 are the feature maps from last three stages of the ResNet-50 network. Enc denotes the output of the final Transformer encoder.}
  \label{tab2}
\end{table} 

\begin{table}[t]
  \centering
  \scalebox{0.75}{
  \begin{tabular}{@{}c cccccc@{}}
    \toprule
      Prompt && HOTA$\uparrow$ & AssA$\uparrow$ & MOTA$\uparrow$ & IDF1$\uparrow$ & IDS$\downarrow$ \\
    \midrule
      Textual prompt          && 16.1 & 17.9 & 10.4 & 19.2 & 6434 \\
      Visual prompt            && 17.5 & 20.4 & 13.7 & 21.4 & 5138\\
    \midrule
      \textbf{Both}   && \bf 19.2 & \bf 21.9 & \bf 16.7 & \bf 24.1 & \bf 4674 \\
    \bottomrule
  \end{tabular}}
  \caption{Comparisons between using different prompts.}
  \label{tab3}
\end{table}  

\begin{table}[t]
  \centering
  \scalebox{0.765}{
\small
  \begin{tabular}{@{}c cccccc@{}}
    \toprule
      Adapter && HOTA$\uparrow$ & AssA$\uparrow$ & MOTA$\uparrow$ & IDF1$\uparrow$ & IDS$\downarrow$ \\
    \midrule
      w          && \bf 19.2 & \bf 21.9 & \bf 16.7 & \bf 24.1 & \bf 4674\\
      w/o            && 16.7 & 16.9 & 14.8 & 19.9 & 6835\\
    \bottomrule
  \end{tabular}}
  \caption{Comparisions between VLM with (w) and without (w/o) the adapter.}
  \label{tab4}
\end{table}

As shown in Tab. \ref{tab2}, the output from the Transformer encoder leads to the best performance, and using the c4 feature behaves better than employing the c3 and c5 features. This phenomenon reveals that both the resolution and semantic information of the visual features are important for generating visual prompts. Due to the deformable attention \cite{zhu2020deformdetr}, the final output of the encoder contains multi-scale semantic context knowledge. Therefore, we choose the output of the transformer encoder as the input to VCP.

\noindent \textbf{Analysis of the visual and textual prompts.} As mentioned before, textual and visual prompts are obtained by Trackbook and VCP, respectively, then they are encoded as PTD in VLM. In this part, we verify the performance of only using one of them, and the results are reported in Tab. \ref{tab3}. As shown, both the textual and visual prompts contribute to the final tracking performance. When we combine both the text information from textual prompts and the image information from visual prompts to produce PTD, the best result is obtained. This observation suggests that incorporating textual description information is helpful for improving MOT performance, and combining the knowledge of multiple modals is a promising research direction.

\noindent \textbf{Analysis on the adapter.} In this part, we analyze the importance of the adapter in the proposed VLM module. The tracking performance corresponding to the models with and without the adapter is reported in Tab. \ref{tab4}. We can observe that the model without the adapter behaves significantly poorer. Specifically, by removing the adapter, the metric HOTA is decreased by $2.5\%$ ($19.2\%$ - $16.7\%$) and AssA is decreased by $5.0\%$ ($21.9\%$ - $16.9\%$). This issue suggests that directly using the output of the CLIP text encoder to generate PTD through cross-attention is not proper. We think this is because of the significant difference between textual prompts and visual prompts, and the adapter can alleviate this difference. Therefore, the adapter module is necessary for VLM. 

\subsection{Visualization}

\begin{figure}[t!]
\centering
\includegraphics[width=0.85\columnwidth]{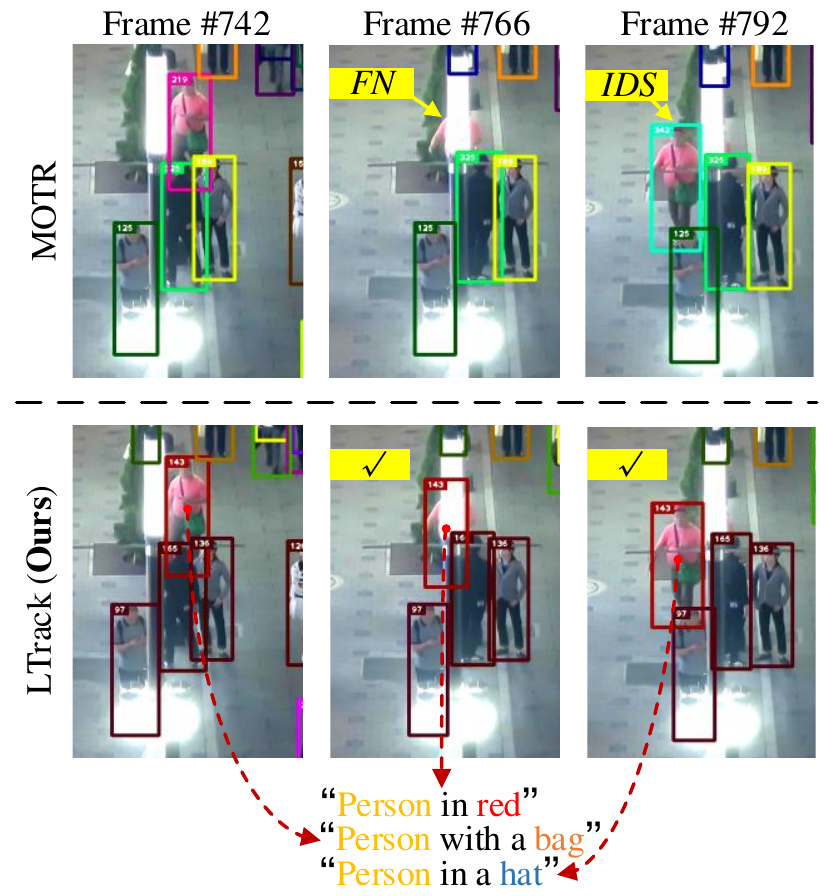} 
\caption{\textbf{Qualitative comparison between the LTrack and the baseline method, MOTR.} \textbf{FN}: False negative, \textbf{IDS}: Identity switch.}
\label{fig5}
\end{figure}

To better demonstrate the superiority of LTrack, we visualize some tracking cases of our method and the baseline, MOTR. One case is illustrated in Fig. \ref{fig5} and it is selected from the testing set of MOT17. As shown, a person is partly occluded and hard to track. The original track query in MOTR can not provide clear guidance to networks on how to track the target correctly, which results in the tracklet disconnection (FN) and identity switch (IDS). On the contrary, LTrack recognizes targets and generates trajectories successfully by using the track queries with textual description information.  This result further confirms the value of PTD in LTrack, which guides the tracker to focus on targets effectively. 

\subsection{More Result and Discussion}

\noindent \textbf{Performance on BDD100K.} We further evaluate LTrack on a autonomous driving dataset, BDD100K \cite{yu2020bdd100k}. As shown in Tab. \ref{tab5}, LTrack still achieves better performance than MOTR on BDD100K. However, the improvement is not that remarkable. We speculate this phenomenon is due to the lack of text descriptions about autonomous driving scenes. This issue merits further exploration in future work.

\begin{table}[h!]
  \centering
  \scalebox{0.765}{
  \begin{tabular}{c ccccc@{}}
    \toprule
      Method && Data & HOTA-pedestrian$\uparrow$ & AssA$\uparrow$ & IDF1-pedestrian$\uparrow$ \\
    \midrule
      MOTR          && CH+17 & 32.4 & 38.8 & 39.3\\
      LTrack            && CH+17 & \bf 33.7 & \bf 39.3 & \bf 40.6\\
    \bottomrule
  \end{tabular}}
  \caption{Performance on BDD100K (pedestrian only).}
  \label{tab5}
\end{table}

\noindent \textbf{Limitation and future work.} LTrack makes a first attempt for the cross-domain MOT task and achieves promising performance, but it still leaves some issues deserving further study. For example, LTrack does not behave well when many similar targets are shared by the same textual information appearing in the same frame. Thus, leveraging more fine-grained textual descriptions may be a future direction to resolve this long-standing problem in MOT. Another limitation is the related open-vocabulary MOT dataset. The paper explains that LTrack only uses 56 coarse-grained textual descriptions in Trackbook. If we have fine-grained textual captions for every target in the video, we can use textual knowledge to guide the tracker explicitly. Nevertheless, the labeling process is very expensive. Hence, to further explore the problem of how to use textual information to guide MOT models, an open-vocabulary MOT dataset is needed.

\section{Conclusion}
In this work, we have pointed out that the domain generalization ability of MOT is hardly studied. To bridge this gap, we first highlighted that the knowledge contained in natural language is inherently more invariant to various domains. Based on this insight, we have implemented a new tracker, namely LTrack, which contains VCP and VLM module to combine the textual and visual prompts. We hope this work can shed light on how to develop MOT trackers with promising generalization ability to some extent.


\appendix
\section{Appendix}

In this file, we provide more details about the proposed method (LTrack) for multi-object tracking (MOT) due to the 7-pages limitation on paper length.

\subsection{CLIP Text Encoder.}  
\begin{figure}[htbp]
\centering
\includegraphics[width=1.0\columnwidth]{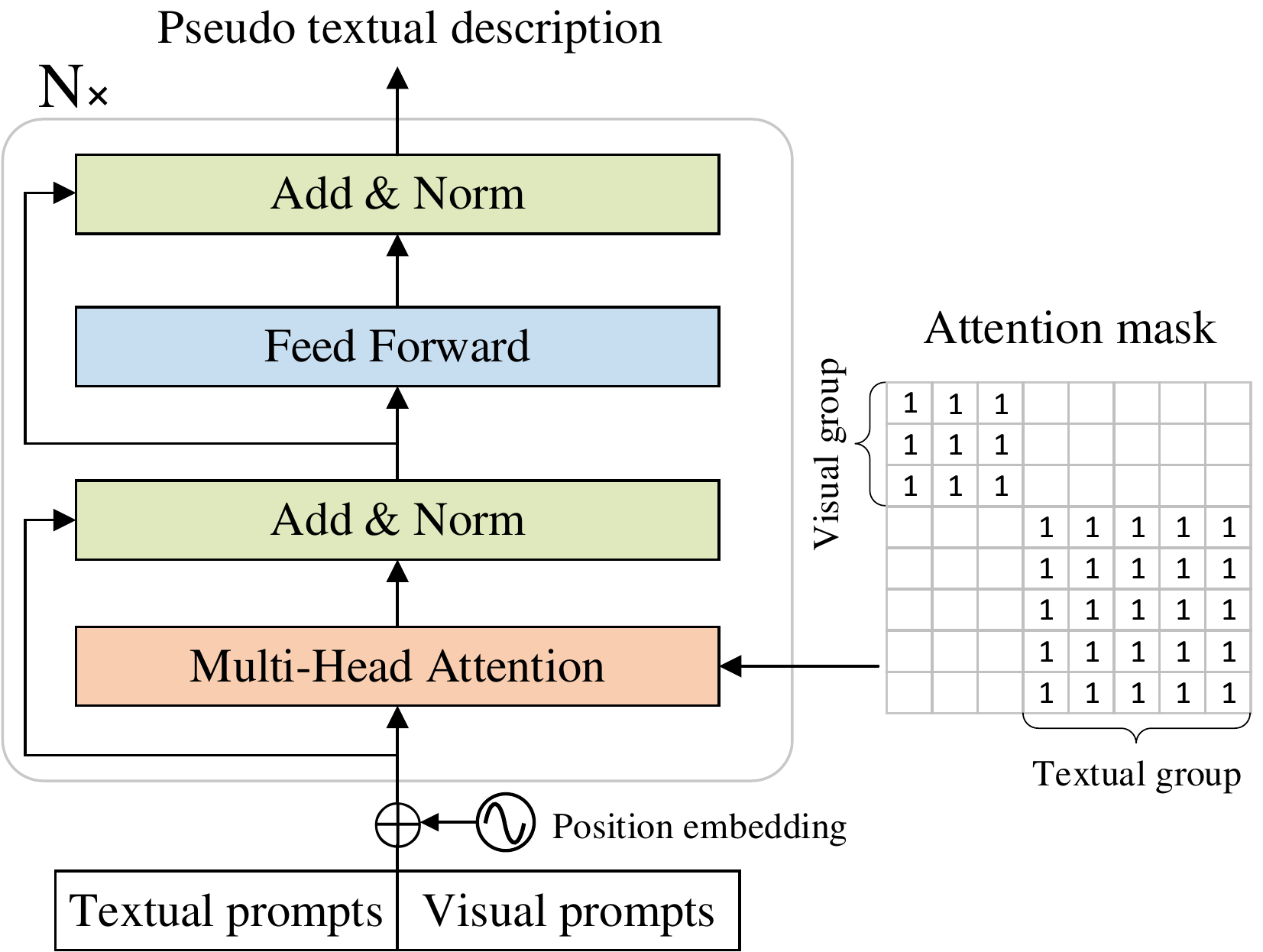} 
\caption{\textbf{Illustration of the masked attention.} There are two parts of input, the textual prompts and visual prompts. The attention masks corresponding to the interaction within textual and visual features are set to $1$. The other mask values are set to $0$.}
\label{fig:mask}
\end{figure}

As introduced in the section of VLM, we adopt the text encoder of CLIP\cite{radford2021learning} to generate pseudo textual description. Since both the visual and textual prompts should be processed by the CLIP text encoder independently, directly passing them to the CLIP text encoder separately consumes double forward time. To alleviate this problem, we design a simple but effective strategy, namely masked attention. Specifically, we concatenate the visual and textual prompts as a single embedding and input this embedding to the CLIP text encoder simultaneously. In order to avoid the information leakage problem between the visual and textual prompts, we build an attention mask as illustrated in Fig. \ref{fig:mask}. This attention mask guarantees that the visual prompt representation does not interact with the textual prompt representation during the attention process, and we can divide them from the output of VLM.

\subsection{Trackbook} 
\begin{figure*}[h!]
\centering 
\includegraphics[width=2.0\columnwidth]{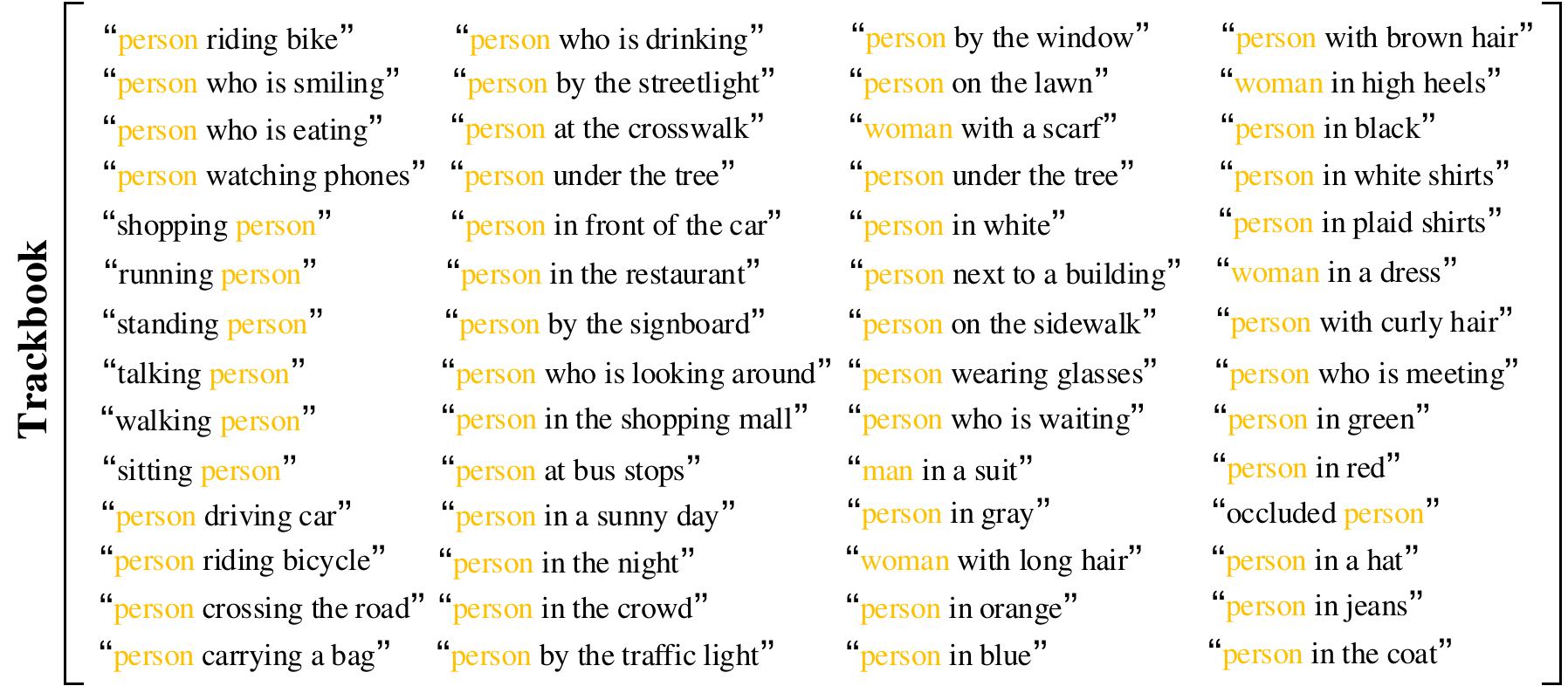} 
\caption{Textual description of targets in Trackbook.}
\label{trackbook}
\end{figure*}

As mentioned before, it is infeasible to label every target with a description sentence. Therefore, we create a Trackbook, which comprises 56 description phrases. Our experiments show that these phrases are enough to cover most tracking cases. The phrases in this Trackbook are presented in Fig. \ref{trackbook}. Our proposed method will convert the description phrases as textual prompts and fuse them with visual prompts to boost the generalization performance of query-based trackers.

\subsection{Additional Experiments}

The effectiveness of various components in our LTrack has been confirmed by the extensive experiments in the manuscript. In this section, we present some additional experimental results to further analyze our method.

\subsubsection{Analysis of the textual prompts.}

As introduced in Section of CLIP, we construct a set of textual prompts based on a fixed template. Previous works have verified that the prompt template significantly affected the performance of the CLIP model. In this part, we study how prompt template affects the tracking performance. The results are reported in Tab. \ref{tab:prompts}. We can observe that the prompt template "a photo of $[CLS]$" achieves the best performance on the metric of HOTA, AssA, IDF1 and IDS, which proves the importance of selecting proper prompt templates for joint visual-language modeling.  

\begin{table}[htbp]
  \centering
\small
  \setlength{\tabcolsep}{3pt}
  \begin{tabular}{@{}c cccccc@{}}
    \toprule
	   Prompt && HOTA$\uparrow$ & AssA$\uparrow$ & MOTA$\uparrow$ & IDF1$\uparrow$ & IDS$\downarrow$ \\
    \midrule
	  -          && 19.1 & 21.1 & \bf 18.7 & 23.6 & 5653 \\
      a            && 17.0 & 21.0 & 14.5 & 19.8 & 8175\\
	  A photo of   && \bf 19.2 & \bf 21.9 & 16.7 & \bf 24.1 & \bf 4674 \\
    \bottomrule
  \end{tabular}
  \caption{Comparison between different prompt templates.}
  \label{tab:prompts}
\end{table} 

\subsubsection{Impact of the text token representation length.} 

\begin{figure}[t]
\centering
\includegraphics[width=1.0\columnwidth]{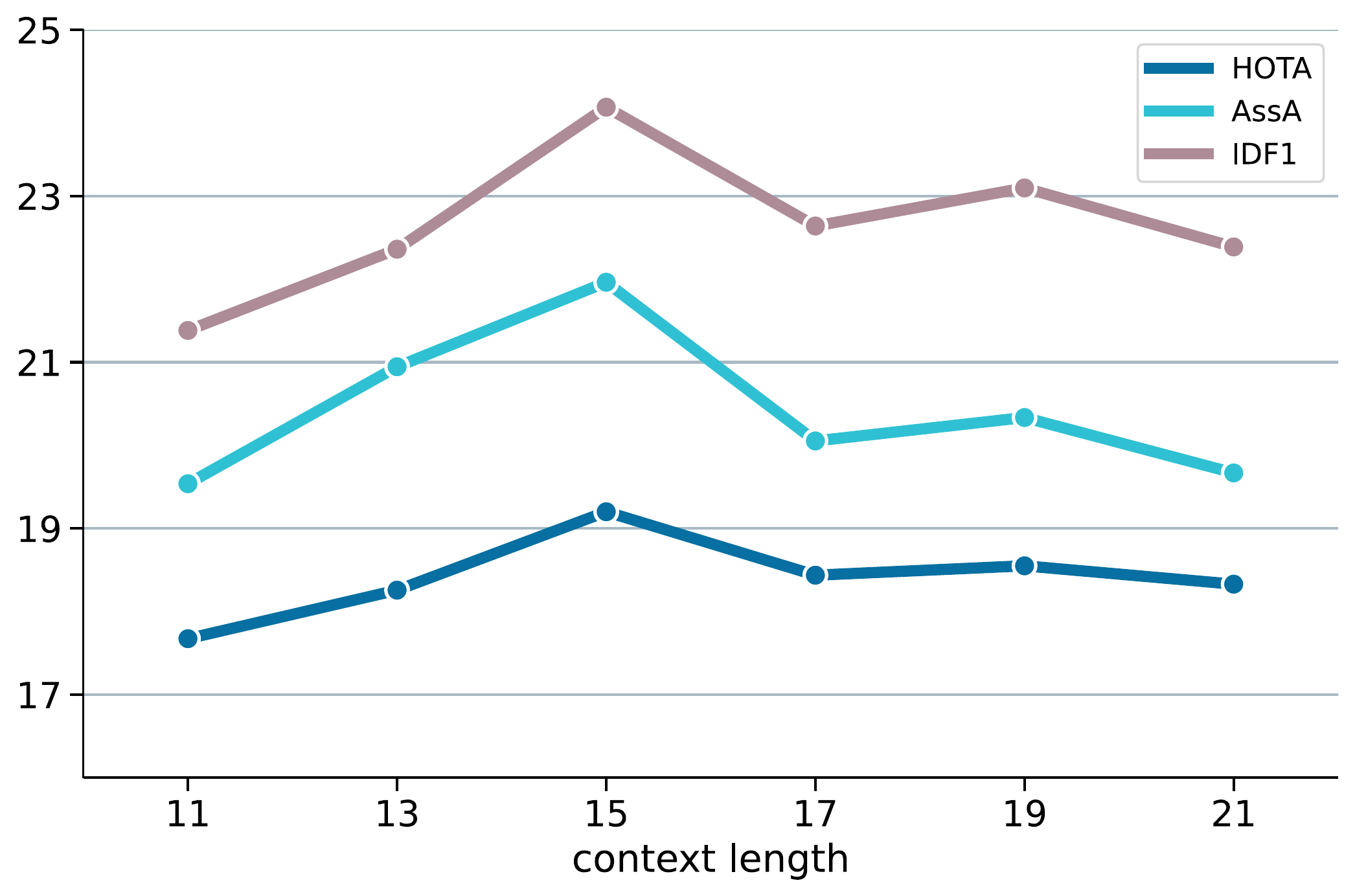} 
\caption{\textbf{Impact of the length of the text token.} The experiment is conducted on our cross-domain benchmark. We mainly compared the metrics of HOTA, AssA and IDF1.}
\label{fig:length}
\end{figure}

The textual prompts are first converted as text token representation before inputting into the CLIP text encoder. In this part, we analyze how the text token length influences the performance during training. The experiment is conducted on the cross-domain benchmark, and the length $L$ is set as 11, 13, 15, 17, 19, and 21, respectively. The results are illustrated in Fig. \ref{fig:length}.

It can be observed from Fig. 7 that the length of text token affects the metrics of HOTA, AssA and IDF1 significantly. When the $L$ is too small or large, the performance of the obtained language embedding is not satisfying, and the best performance is obtained when $L$ is $15$. We speculate that the optimal length is affected by the balance between valuable and reluctant information in this token. When $L$ is too small, the text token cannot contain all the valuable textual information. Meanwhile, By contrast, when $L$ is too large, much useless information is contained, which increases the optimization difficulty.

\subsection{Additional Visualization}

\begin{figure*}[h!]
\centering 
\includegraphics[width=2.0\columnwidth]{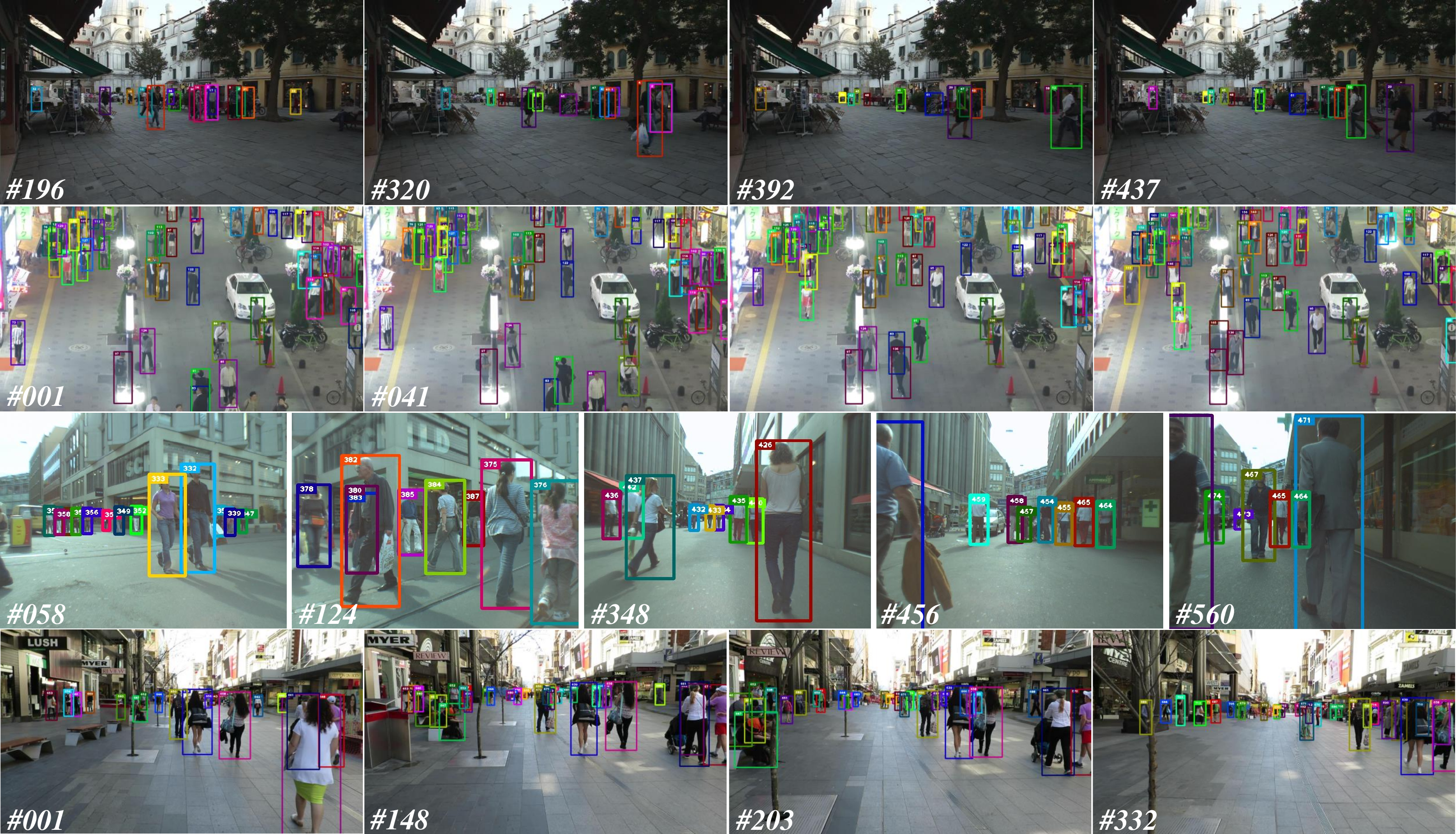} 
\caption{Tracking results of LTrack on the in-domain dataset (MOT17).}
\label{in-domain}
\end{figure*}

\begin{figure*}[h!]
\centering 
\includegraphics[width=2.0\columnwidth]{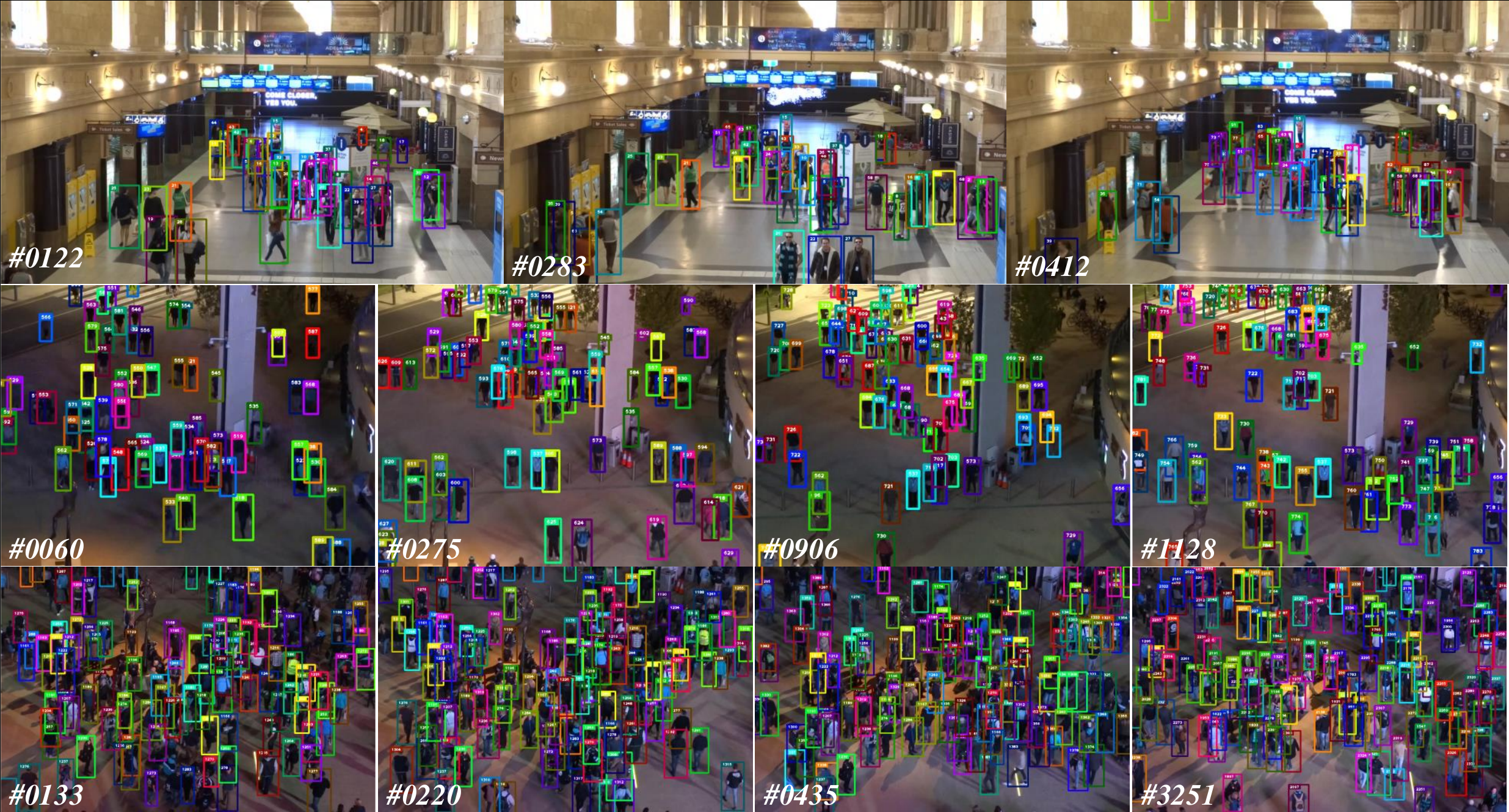} 
\caption{Tracking results of LTrack on the out-domain dataset (MOT20).}
\label{out-domain}
\end{figure*}

More challenging cases are visualized and presented in Fig. \ref{in-domain} and Fig. \ref{out-domain} to further confirm the superiority and generalization of LTrack. The shown cases are chosen from the test splits of the in-domain (MOT17) and out-domain (MOT20) datasets. As shown in Fig. \ref{in-domain} and Fig. \ref{out-domain}, we can observe that there is a huge gap between in-domain and out-domain scenes, including indoor and outdoor scenarios, light and dark brightness, huge and small targets, etc. And our LTrack achieves excellent tracking performance in both in-domain and out-domain scenes. 

\subsection{Limitation and Future work}

\subsubsection{Limitation.} Although LTrack achieves remarkable performance on the cross-domain MOT benchmark, there are still some issues deserving further study. Specifically, LTrack does not behave well when many targets with similar appearance and actions occur in the same frame. This is because it is difficult to distinguish the targets using the textual information in the description phrase or semantic information in track queries. For this kind of scenes with many similar targets, the appearance information is also not reliable, and we can only seek help from spatial priors or complex post-processing.

Another limitation of this work is the lack of large open-vocabulary MOT datasets. As explained in the paper, LTrack only uses the $56$ coarse-grained textual descriptions in Trackbook. If we have fine-grained textual caption for every target in the video, we can use textual knowledge to guide the tracker with explicit supervision. Nevertheless, the labeling process is very expensive and no such kind of datasets emerges so far.

\subsubsection{Future Work.} According to the above analysis of the limitation, there are two promising research topics for our future works.

(1) How to distinguish similar targets during tracking is a long-standing problem in MOT task. Leveraging more fine-grained textual description is a possible option.

(2) The process of collecting and annotating text labels is expensive. Therefore, it is valuable to study how to get data from the Internet and train trackers in a unsupervised way.

{\small
\bibliography{aaai23}
}
\clearpage
\end{document}